\documentclass{article}
\usepackage{spconf,amsmath,graphicx}

\usepackage{amsfonts}
\usepackage{booktabs}

\title{TCT: A Cross-supervised Learning Method for Multimodal Sequence Representation}
\name{Wubo Li, Wei Zou, Xiangang Li}
\address{Didi Chuxing, Beijing, China\\
\{liwubo, zouwei, lixiangang\}@didiglobal.com}
\begin{document}
\ninept
\maketitle
\begin{abstract}
Multimodalities provide promising performance than unimodality in most tasks. However, learning the semantic of the representations from multimodalities efficiently is extremely challenging. To tackle this, we propose the Transformer based Cross-modal Translator (TCT) to learn unimodal sequence representations by translating from other related multimodal sequences on a supervised learning method. Combined TCT with Multimodal Transformer Network (MTN), we evaluate MTN-TCT on the video-grounded dialogue which uses multimodality. The proposed method reports new state-of-the-art performance on video-grounded dialogue which indicates representations learned by TCT are more semantics compared to directly use unimodality.
\end{abstract}
\section{Introduction}
\label{sec:intro}

The world surrounding us includes multifarious modalities to express messages. To make progress in the understanding world, we need to be able to interpret and reason about multimodal messages. Multimodal machine learning aims to build models that can process and relate information from multiple modalities. The research field of Multimodal Machine Learning brings some unique challenges for computational researchers given the heterogeneity of the data. Learning from multimodal sources offers the possibility of capturing correspondences between modalities and gaining an in-depth understanding of natural phenomena\cite{baltruvsaitis2018multimodal}.

Inspired by the McGurk effect \cite{mcgurk1976hearing}, one of the earliest research on the multimodality is audio-visual speech recognition (AVSR)\cite{yuhas1989integration} which was an interaction between listening and vision during speech recognition. The field of multimedia content indexing and retrieval is the second important category of multimodal learning applications\cite{atrey2010multimodal,snoek2005multimodal}. Multimodal interaction is a straight way to study multimodal behaviors of humans, such as audio-visual emotion recognition\cite{schuller2011avec}. Recently, the attention of multimodal learning is transferred to visual and language for media description and question answering, such as video caption \cite{krishna2017dense}, video question answering\cite{xu2017video} and video dialog \cite{alamri2019audio}.

Most existing methods for multimodal sequence representations use RNN as the sequence processing unit and model the sequence information with a sequence-to-sequence network as the overall architecture. To combine with multimodal features, \cite{DBLP:conf/iccv/HoriHLZHHMS17} proposed to use hierarchical attention to fuse contributions from different modalities. In the  \cite{DBLP:conf/iccv/HoriHLZHHMS17,sanabria2019cmu}, the video and audio are split into the fixed-frame to extracted and using pre-trained I3D, VGG and 3D ResNeXt to compute semantic features. Because of the drawback of the long-term dependence in RNNs, \cite{le2019multimodal} proposed using transformer to encode semantic representations in multimodal dialog systems. Previous works mainly concentrate on expressing different modal features by using pre-trained extractor and attention-based encoder, but ignore using the consistency of the related multimodality sequence to learning the unimodality sequence on a supervised learning method.

In this paper, we proposed the Transformer based Cross-modal Translator (TCT) for unimodality sequence representations. In the TCT,  unimodality sequence representations are translated from other related multimodalities sequences on a supervised learning method. The key to our method is that we effectively make use of correlative multimodalities to expressing target modality by supervised learning. Combined with TCT, we proposed a novel method MTN-TCT for video-grounded dialog implemented on the state-of-the-art model Multimodal Transformer Network (MTN). We evaluated MTN-TCT on the AVSD dataset which uses multimodalities to generate question responses. The proposed approach reports new state-of-the-art performance on video-grounded dialogue. The analysis of MTN-TCT indicates that the modality learned from TCT is much more semantical than directly using unimodality.

\section{Related Work}
\label{sec:related-work}
In order to understand semantics information from the image, video, audio and other modalities, and provide more precision representations for multimodal learning applications. Many works have done on pretraining unimodal feature extractors and using extra information to improve performances. For audio and video features, many computer vision and speech tasks have a similar scene, such as video caption, image caption and sound event detection. Thus, it is easy to investigate transfer learning to achieve semantic representations. The end-to-end audio classification ConvNet and AclNet are incorporated into video-grounded dialog model on \cite{kumar2019context}. Sanabria et al. use 2000 hours of how-to videos \cite{sanabriahow2} to pre-train a video features extractor and regard video-grounded dialog task as video caption to generate answers \cite{sanabria2019cmu}. 

Exploring powerful architecture to encode text and non-textual modalities is another effective method to obtain semantic representations. The work in \cite{yeh2019reactive} uses a simple but effective $1\times1$ convolution to fuse multimodal features and propose a multi-stage fusion mechanism to thoroughly understand the question. Inspired by FiLM, FiLM Attention Hierarchical Recurrent Encoder-Decoder (FA-HERD) \cite{nguyen2019film} is proposed to condition the audio and video features on question to reduce the dimensionality considerably.

The most related work with ours is Multimodal Transformer Networks (MTN) proposed by Le et al. \cite{le2019multimodal}. In MTN, the Recurrent Neural Networks are replaced by the transformer, and query-aware attention encoder is proposed to obtain question-related no-textual features. Different from MTN and FA-HERD, our TCT using other multimodality sequences to translate the target modality sequence by supervised learning method which is more general and task-independent.

\section{Our Approach}
\label{sec:approach}

\subsection{TCT}
\label{ssec:tct}

Transformer based Cross-modal Translation is aimed to learn a target modality sequence from other related multimodal sequences by a supervised method. Given the target modality sequence as $X = {\{x_1, x_2,..., x_T\}}$ and source modality sequence $\widehat{X}^k=\{\widehat{x}^{k}_{1}, \widehat{x}^{k}_{2},..., \widehat{x}^{k}_{N}\}$, where $T$ and $N$ is the sequence length of the target modality and source modality, $k=\{1,2,..., K\}$ indicates different modalities. In theory, TCT can learn the target modality sequence from K modalities sequences. For easy to explain, here, we show an example to explain how to translate target modality sequence $X$ from $k_{th}$ source modality sequence. For the complex case, we can concatenate the multimodality sequence as a large fusion sequence to translate target modality.

\begin{figure}[htb]
  \centering
  \centerline{\includegraphics[width=8.5cm]{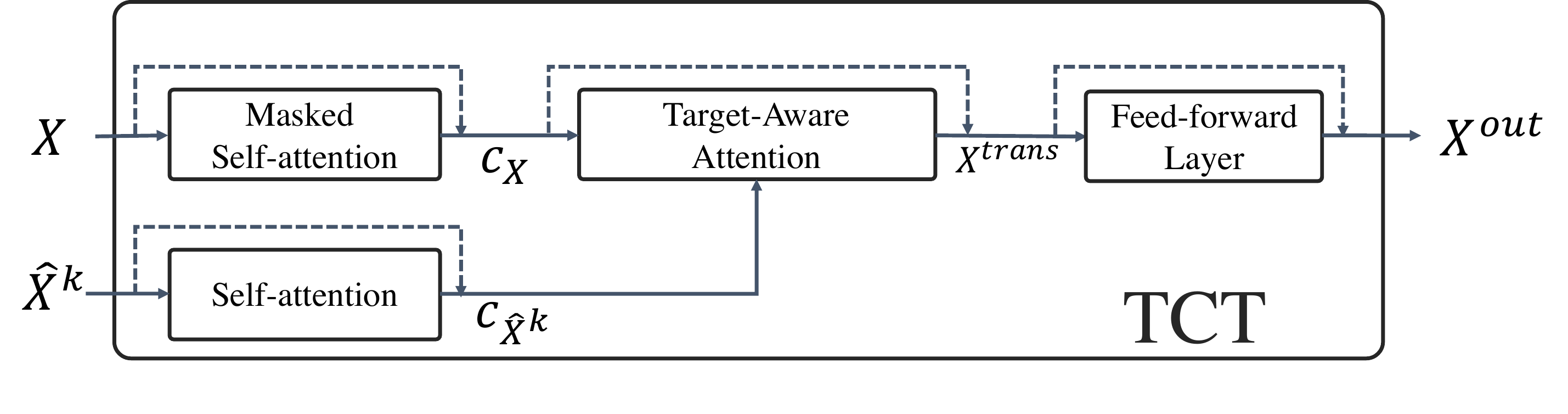}}
  \vspace{-1.0cm}
  \centerline{}\medskip
\caption{The basic block of TCT.}
\label{tct}
\end{figure}

The basic block of the TCT mainly consists of three components: two self-attention and an attention mechanism which regards target modality sequence as query and another modality sequence as key and value \cite{vaswani2017attention}. The target modality sequence $X$ and $k_{th}$ modality sequence $\widehat{X}^k$ are fed into masked self-attention and self-attention respectively. The output sequences of the first two self-attention is $c_X$ and $c_{\widehat{X}^{k}}$. To translate target modality $X$ by modality $\widehat{X}^k$, an attention mechanism is used as a translator and the output is denoted as $X^{trans}$. At the last of the TCT block, we employee two feedforward layers that have parallels with \cite{vaswani2017attention} and the output sequence is denoted as $X^{out}$. We employ residual connection and layer normalization around the attention and feed-forward layer. The general attention and feed-forward layer computations mentioned above are as follow:
\begin{normalsize}
\begin{equation*} \label{eqn2}
  \begin{split}
  &o = max(0, mW_1 + b_1)W_2 + b_2 \\
  &m = Concat(h_1,...,h_h)W^O    \\
  &h_i = Attn(QW^{Q}_i,KW^{K}_i,VW^{V}_i) \\ 
  &Attn(Q,K,V) = softmax(\frac{QK^T}{\sqrt{d_k}})V \\
  \end{split}
\end{equation*}
\end{normalsize}
where $h$ is the number of heads and $d$ is the demension of attention, $W^Q_{i},  W^K_{i}, W^V_{i} \in \mathbb{R}^{d \times d}$, $W^O \in \mathbb{R}^{hd_v \times d}$. The goal of TCT is predicting target modality sequence $X$ by the output sequence $X_{k}^{out}$. For supervised learning, TCT employs log-likelihood function for textual modality sequence and $L1$ loss or similarity loss \cite{kovaleva-etal-2018-similarity} for the other dense modalities sequence.

\subsection{MTN-TCT}
\label{ssec:mtn-tct}
\begin{figure*}[ht]
  \centering
  \centerline{\includegraphics[scale=0.51]{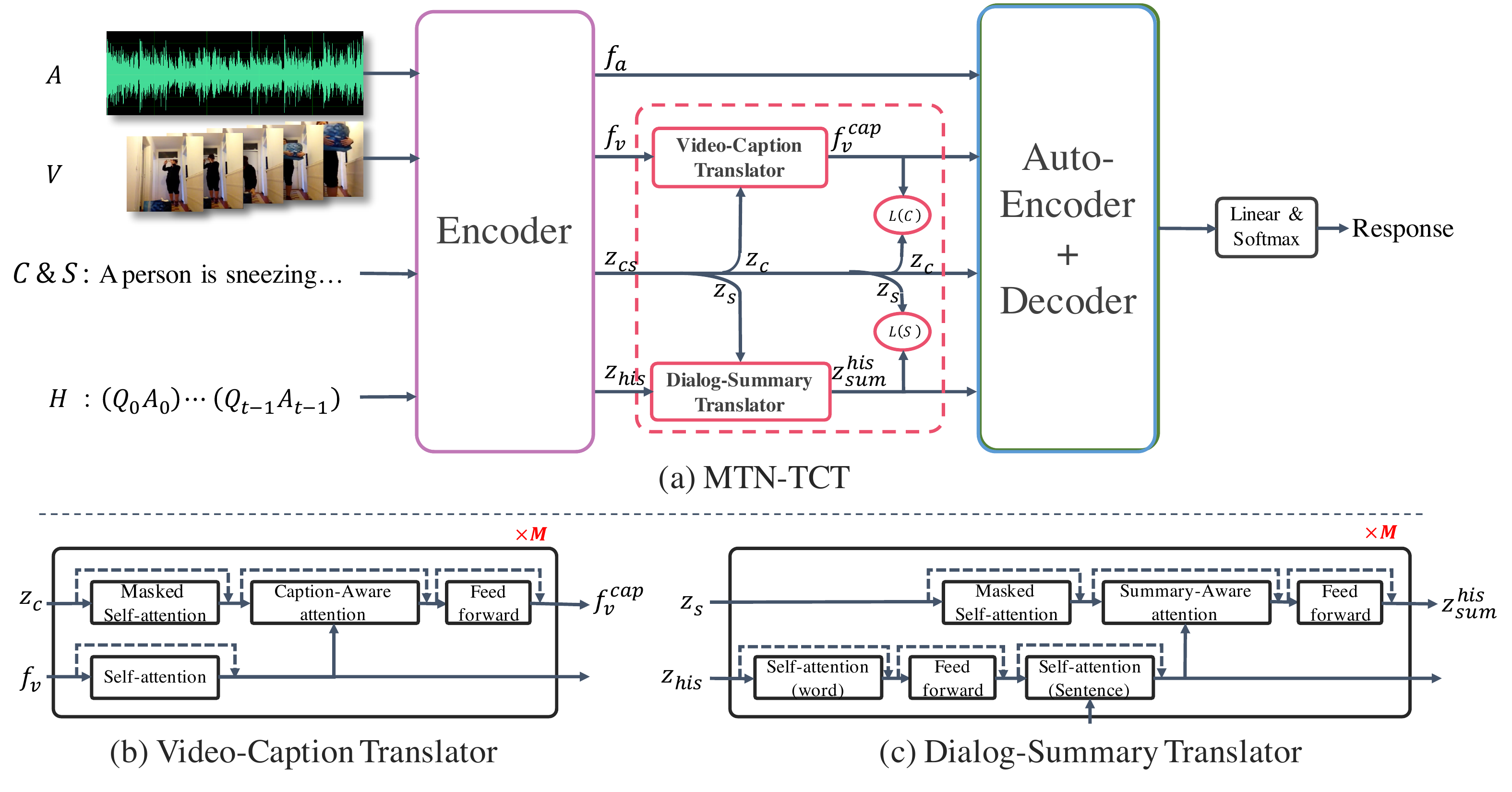}}
  \vspace{-0.7cm}
  \hspace{-6.0cm}
  \centerline{}\medskip
\caption{(a) Overall of the MTN-TCT. The red dotted rectangle indicates Video-Caption Translator and Dialog-Summary Translator. For simplicity, we have omitted the input for questions and answers in overall of model. (b) Video-Caption Translator. (c) Dialog-Summary Translator.}
\label{overall}
\end{figure*}

Our MTN-TCT is based on the Multimodal Transformer Networks proposed in \cite{le2019multimodal} which consists of five modules: Encoder module, Decoder module, Auto-Encoder module, Video-Caption Translator module and Dialogue-Summary Translator module. In the Encoder module, the text sequence and video features are mapped to a sequence of continuous representations $ z=(z_1,...,z_n) \in \mathbb{R}^{d} $, $ f_a $ and $ f_v $. An Overview of text sequence and video features encoder can be seen in Figure \ref{overall}. Video-Caption Translator is $M$ blocks of TCT that allow the model to learn visual features well. MeanWhile, the Dialogue-Summary Translator module read the history of the dialogue, and learn to summary the QA pairs to provide semantic representations. The essential of the Auto-Encoder module is applied for the interaction between the query and non-text modalities. In this module, the attention to allow the model to focus on query-related features of the video. We follow the architecture in \cite{le2019multimodal}, decoder module is consist of stack transformer layer to fusing all modal features and generate an output sequence $y=(y_1,...,y_m)$ including a start-of-sentence token (sos) and an end-of-sentence token(eos).  Next, we will briefly introduce the Video-Caption Translator module and Dialogue-Summary Translator.

\subsubsection{Video-Caption Translator } \label{subsec:vc-translator}
The input sequences of Video-Caption Translator are the visual representations $ f_v $ learned from the Encoder module. To sufficient learn the semantics representation of the video, we proposed a $M$ blocks of TCT to regenerate the caption as shown in Figure \ref{overall}(b). The visual feature representations $f_v$ as source modality and the tokenized caption $z_c$ as target modality are feed into blocks of TCT. The output of the Video-Caption translator module are caption-related visual representations $f^{cap}_v$.

\subsubsection{Dialogue-Summary Translator} \label{subsec:ds-translator}
We proposed a hierarchical TCT architecture to encode dialogue history and regenerate a summary as shown in Figure \ref{overall}(c). The difference hierarchical TCT with TCT is that we encode source modality in two-levels. For example, in the Dialogue-Summary Translator, tokenized dialogue history sequences $z_{his}$ encoded by a single transformer layer to generate word-level contextual representations, and then extract eos representations as sentence-level representations of the dialogue. Then summary representations $z_s$ and the sentence-level dialogue representations are fed into TCT to generate summary-related dialogue representations $z^{his}_{sum}$.

\subsubsection{Loss} \label{subsec:loss}
Given the dialogue history ($H$), question ($Q$), video features ($V$ and $A$), video caption ($C$) and summary ($S$), we use the log-likelihood as the objective function for target sequences answer ($Ans$), Video-Caption Translator output ($f^{cap}_v$) and Dialogue-Summary Translator output ($z^{his}_{sum}$) while training MTN-TCT.  The log-likelihood function consists of three parts as follow:
\begin{normalsize}
\begin{equation} \label{eqn1}
  \begin{split}
  L=&L(Ans) +\alpha L(C) + \beta L(S)    \\
  =&logP(Ans| H, Q, V, A, C, S) + \alpha logP(C| V)  + \\
  & \beta logP(S| H).
  \end{split}
\end{equation}
\end{normalsize}
where we fix $\alpha=1$ and $\beta=1$ while training MTN-TCT in the video-grounded dialog.

\section{Experiment}
\label{sec:experiment}

\subsection{Data}
\label{ssec:data}	
We evaluated MTN-TCT on an Audio Visual Scene-aware Dialog (AVSD) dataset proposed on \cite{Hori_2019}. The AVSD dataset consists of Q\&A conversations about short videos obtained from two Amazon Mechanical Turk (AMT) workers, who discuss events in a video. In each dialog, one of the workers takes the role of an answerer who had already watched the video. The answerer replies to questions asked by another AMT worker, the questioner. The dataset contains 9,848 videos taken from CHARADES, a multi-action dataset with 157 action categories \cite{Sigurdsson_2016}.

\begin{table}[ht]
\centering	
        \caption{DSTC7 Audio Visual Scene-aware Dialogue Dataset}
	\begin{tabular}{lccc}
    \hline
	& Train & Validation & Test  \\ \hline
	num of Dialogs     & 7,659          & 1,787               & 1,710          \\ 
	num of Turns       & 153,180        & 35,740              & 13,490         \\ 
	num of Words       & 1,450,754      & 339,006             & 110,252        \\ \hline
	\end{tabular}
	\label{AVSD_data}
	\end{table}

We used 7,659 dialogs for the training set and 1787 for dev set, and evaluated our model on the official test set which contains 1710 dialogs \cite{Hori_2019}. Table \ref{AVSD_data} summarizes the dataset.
	
In the AVSD7, the metrics are commonly used in natural language process tasks, such as BLEU(1-4) \cite{papineni2002bleu}, METEOR \cite{denkowski2014meteor}, ROUGE-L \cite{lin2004automatic} and CIDEr \cite{vedantam2015cider}.

\subsection{Training}
\label{ssec:training}
Our MTN-TCT model is based on MTN which is the state-of-the-art on AVSD dataset. Following the MTN, the dimension of the query, key and value in MTN-TCT is 512, the number of the attention heads is 16. The MTN-TCT consists of 6 layers Decoder and Auto-Encoder. We use dropout probability is 0.5, and warm up the schedule with ${warmup\_step}$ 13000 for modifying the learning rate while training. Adam optimizer \cite{kingma2014adam} is used for training MTN-TCT. The number of TCT blocks used in Video-Caption Translator and Dialog-Summary Translator ($M$ mentioned in  Figure \ref{overall}) is 1. For all models, we evaluated on the test set of the AVSD by loading the lowest perplexity on the validation set.

\subsection{Results}
\label{ssec:results}
We compared MTN-TCT with the Hierarchical Attention Network (Naive) proposed on \cite{Hori_2019} and current state-of-the-art model MTN \cite{le2019multimodal}. In the AVSD track of DSTC7, as shown in Table \ref{avsd_result}, the models can be evaluated on 1 reference and 6 references. Row 1 to 3 show the HAN, MTN and MTN-TCT with 1 reference. While referencing on 6 candidates, as shown in row 4 to 6 of Table \ref{avsd_result}, MTN-TCT outperforms the all model in terms of METEOR, ROUGE-L and CIDEr. However, BLUE scores are a little poor compared with MTN. The higher ROUGE-L indicates that MTN-TCT prefers to generate longer responses to the questioner. We deduce that due to sufficient translation from the multimodal sequences, MTN-TCT can describe more details in video-grounded dialog. Meanwhile, the higher scores of the METEOR and CIDEr show that the responses of the MTN-TCT are more similar to humans. 

\begin{table}[ht]
	\renewcommand\arraystretch{1.5}
	\centering
	\caption{Automatic evaluation metrics of DSTC7 Audio Visual Scene-aware Dialogue Dataset. The last three models are evaluated on official 6 references.}
	\resizebox{0.49\textwidth}{!}{ 
	\setlength{\tabcolsep}{0.5mm}
	\small
	\begin{tabular}{clccccccc}
   \toprule
	No. & Description & BL-1 & BL-2 & BL-3 & BL-4 & METEOR & ROUGE-L & CIDEr \\  \hline
	\multicolumn{9}{c}{Official 1 reference} \\\hline
	 1 & HAN (Naive) \cite{Hori_2019}   & 0.279 & 0.183 & 0.130 &0.095& 0.122& 0.303& 0.905\\ 
	 2 & MTN \cite{le2019multimodal} &0.357 &0.241 &0.173 &0.128 &0.162 &0.355 &1.249 \\ 
         3 & MTN-TCT (Ours)   &\textbf{0.371}& \textbf{0.257}& \textbf{0.185}& \textbf{0.137}& \textbf{0.170}& \textbf{0.368}& \textbf{1.304}\\ \hline
         \multicolumn{9}{c}{Official 6 references} \\   \hline
	4 & HAN (Naive) \cite{Hori_2019} & - & - & - & 0.309 & 0.215 & 0.481 & 0.733\\ 
	5 & MTN \cite{le2019multimodal} &  \textbf{0.723} &  \textbf{0.589} & \textbf{0.480} & \textbf{0.392} & 0.278 & 0.571 & 1.128 \\
	6 & MTN-TCT (Ours) &0.703 &0.571 & 0.467 & 0.382 & \textbf{0.290} & \textbf{0.584} & \textbf{1.162} \\   \toprule
	\end{tabular}
        }
        \label{avsd_result}
	\end{table}

\begin{figure}[htb]
  \centering
  \centerline{\includegraphics[width=7.5cm]{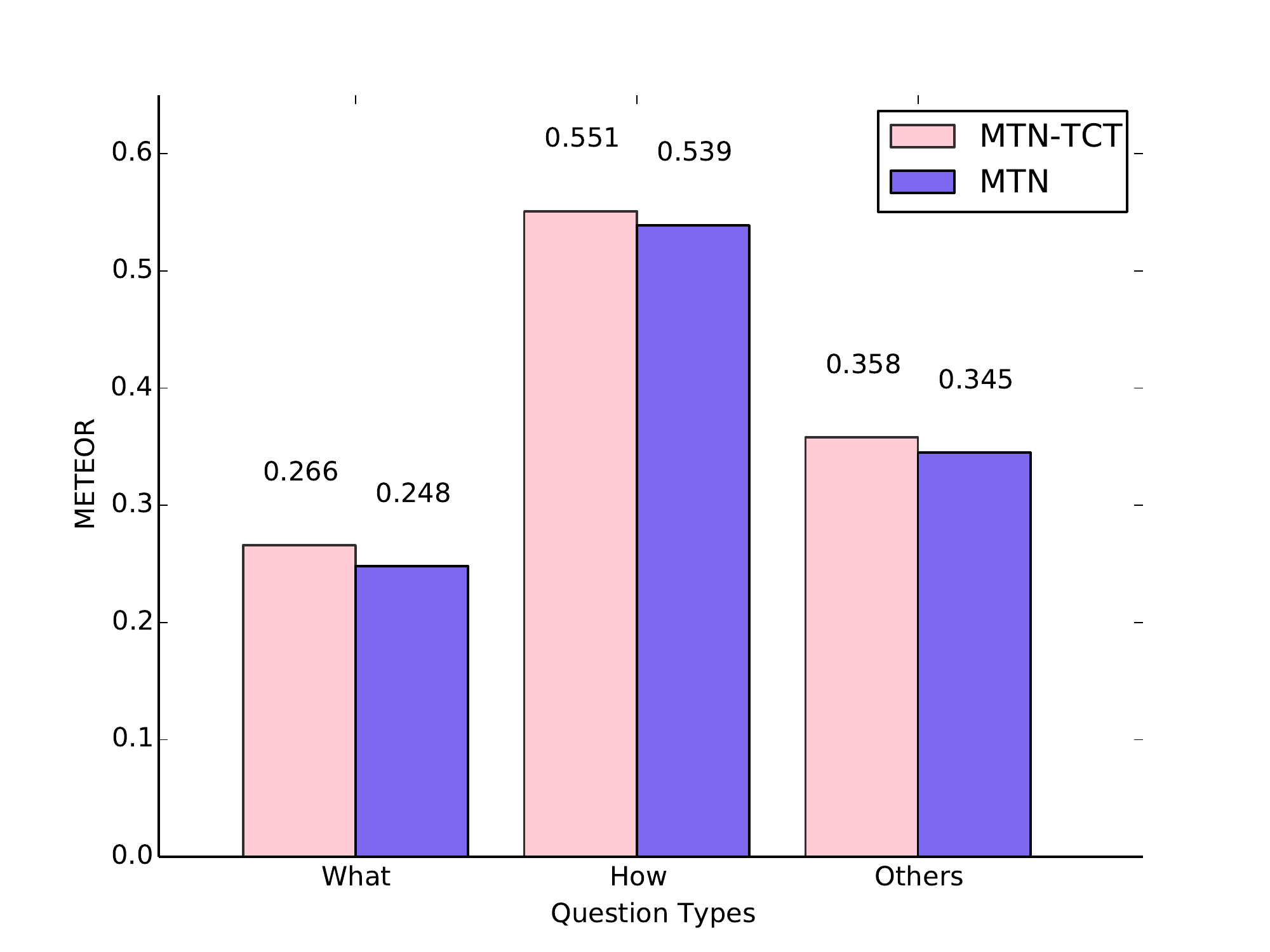}}
  \vspace{-0.6cm}
  \centerline{}\medskip
\caption{Comparisons of MTN and MTN-TCT on Top 3 Type of Questions.}
\label{top3question}
\end{figure}

To explore the performances of MTN and MTN-TCT on different types of questions. We classify the question types into \textit{What}, \textit{Who}, \textit{Where}, \textit{Which}, \textit{How}, \textit{When}, \textit{Why} and \textit{Others}. The top three question types are \textit{What}, \textit{How} and \textit{Others} which make up about $93\%$ of 1710 questions. Figure \ref{top3question} shows METEOR scores (harmonic mean of recall and precision)  on the top three types of questions evaluated on the MTN and MTN-TCT separately.  The results demonstrate that MTN-TCT outperforms the MTN on reasoning questions (why and how questions) \cite{mishra2016survey} which need more details as knowledge. This indicates that the modality sequence learned from TCT is much more semantical than directly using the unimodality sequence.

\begin{figure}[htb]
  \centering
  \centerline{\includegraphics[width=8.5cm]{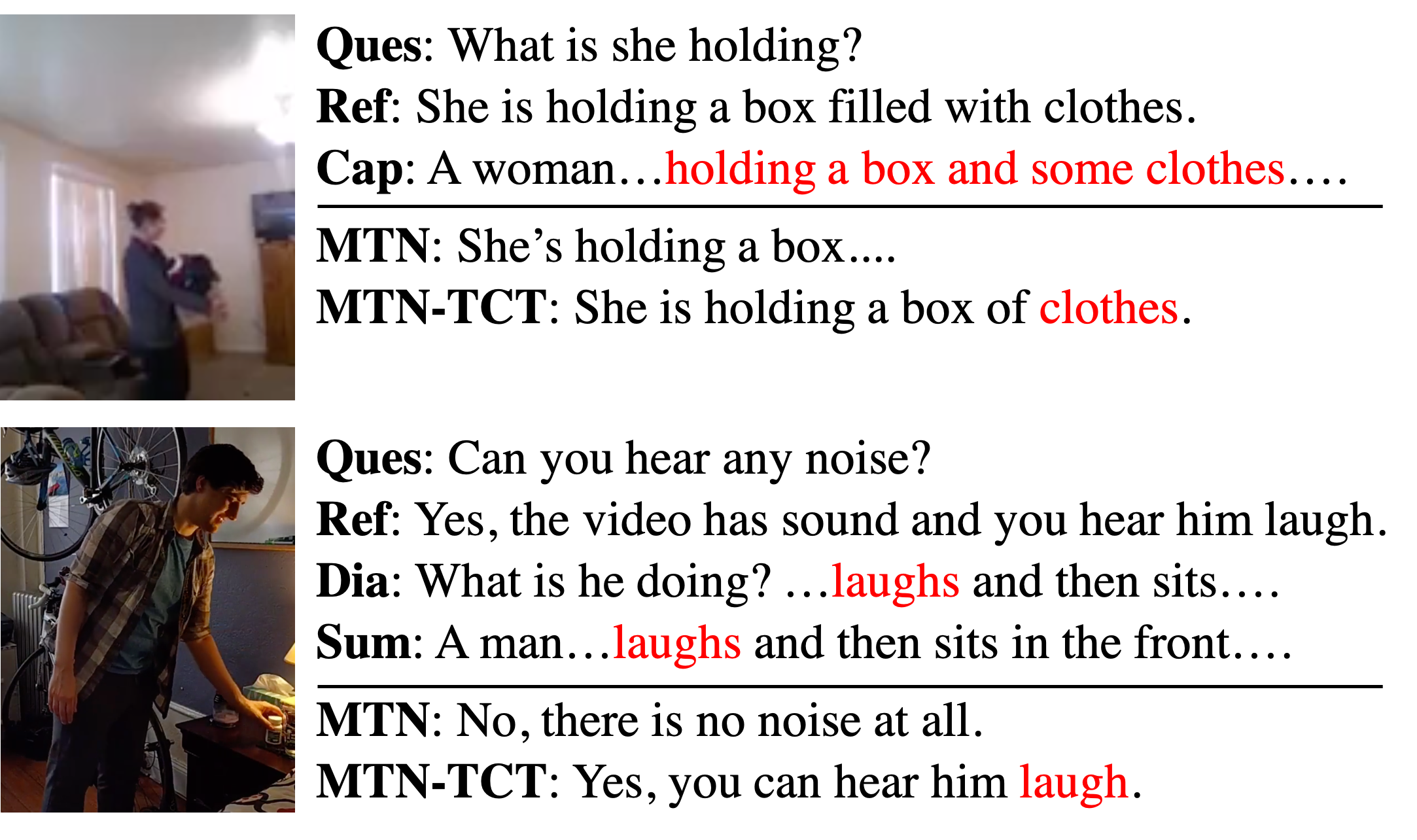}}
  \vspace{-0.6cm}
  \centerline{}\medskip
\caption{Responses examples of MTN and MTN-TCT.}
\label{avsd_examples}
\end{figure}

MTN and MTN-TCT own the same multimodal features on the AVSD dataset, such as video, audio, dialog, caption and summary of dialog. Figure \ref{avsd_examples} shows two response examples generated by MTN and MTN-TCT. In the first example, the question is about what does woman holds in the video. MTN-TCT understands the video frame by translating from the caption and answer ``a box of clothes" (highlight words) instead of ``a box" (response of MTN). In the second video, a man is putting a bottle on the table while laughing. MTN-TCT captures the laughter in dialog history by the summary and gives the response as ``you can hear him laugh" while MTN replies ``no noise at all". The examples demonstrate that MTN-TCT can enhance the ability of understanding unimodality by related multimodality sequences, such as visual-caption in the first example and dialog-summary in the second example.

\subsection{Ablation Study}
\label{ssec:ablation-study}

\begin{table}[ht]
	\renewcommand\arraystretch{1.5}
	\centering
	 \caption{Ablation study in MTN-TCT.}
	\resizebox{0.49\textwidth}{!}{ 
	\setlength{\tabcolsep}{0.5mm}
	\begin{tabular}{clccccccc}
   \toprule
	No. & Description & BL-1 & BL-2 & BL-3 & BL-4 & METEOR & ROUGE-L & CIDEr \\  \hline
	1 & MTN \cite{le2019multimodal} &0.357 &0.241 &0.173 &0.128 &0.162 &0.355 &1.249 \\
	2 & MTN+S & 0.358 & 0.247 & 0.179 & 0.134 & \textbf{0.172} & 0.365 & 1.276\\ 
	3 & MTN+C   & 0.371 & 0.255 & 0.183 & 0.136& 0.170& 0.368& 1.293\\ 
	4 & MTN-TCT (Ours) &\textbf{0.371} &\textbf{0.257} &\textbf{0.185} & \textbf{0.137} & 0.170 & \textbf{0.368} & \textbf{1.304} \\   \toprule
	\end{tabular}
        }
        \label{avsd_result_ablation}
	\end{table}

We conduct an ablation study to verify the influence with the Video-Caption Translator and  Dialogue-Summary Translator in MTN-TCT. For comparison, we remove the Video-Caption Translator and  Dialogue-Summary Translator from MTN-TCT respectively. To eliminate the influence of the transformer depth, we also integrate stacked self-attention blocks into Encoder which does not obtain improvements. Thus, in ablation study, MTN is the base model and MTN-TCT is the model mentioned in row 3 in Table \ref{avsd_result}, MTN+C and MTN+S indicate adding Video-Caption Translator and  Dialogue-Summary Translator in MTN-TCT respectively. As shown in Table \ref{avsd_result_ablation}, the results demonstrate that both Video-Caption Translator and  Dialogue-Summary Translator can help MTN-TCT understanding the semantics of visual modality and textual modality by translating from caption and summary respectively. Row 2 and 3 of Table \ref{avsd_result_ablation} shows that Video-Caption Translator has more benefits than Dialogue-Summary Translator. This phenomenon occurs due to the summary is concluded by the questioner who did not watch the video. It causes the information on the summary is less than the caption, thus the improvement of the Dialogue-Summary Translator is less than Video-caption Translator.




\section{Conclusion}
In this paper, we propose the Transformer based Cross-modal Translator(TCT) to learn unimodal sequence representations by translating from other related multimodal sequences in a supervised learning method. Besides, we proposed MTN-TCT which integrates TCT on textual and visual modal features to learn semantics representations for video-grounded dialog systems. On the AVSD track of the $7^{th}$ Dialog State Tracking Challenge, MTN-TCT reports new state-of-the-art performance compared to MTN and other submission models. The automatic evaluation scores and case analysis demonstrate that MTN-TCT can learn more details from multimodalities sequence. The comparisons with MTN and ablation study indicate that the modality learned by TCT is much more semantics than directly using unimodality.  
\label{sec:conclusion}

\bibliographystyle{IEEEbib}
\bibliography{strings,refs}

\end{document}